\documentclass{article}

\usepackage[table]{xcolor}
\usepackage[preprint]{corl_2026} 
\usepackage{graphicx}
\usepackage{amsmath}
\usepackage{booktabs}
\usepackage{multirow}
\usepackage{wrapfig}
\usepackage{caption}

\usepackage{pgfplots}
\pgfplotsset{compat=1.18}

\definecolor{groupone}{RGB}{102,153,204}  
\definecolor{grouptwo}{RGB}{120,185,130}  
\newcommand{\bestgroupone}[1]{\cellcolor{groupone!20}#1}   
\newcommand{\bestgrouptwo}[1]{\cellcolor{grouptwo!25}#1} 

\usepackage{tikz}
\usetikzlibrary{arrows.meta}
\usetikzlibrary{calc}
\usetikzlibrary{svg.path} 
\usepgfplotslibrary{groupplots}

\title{RayViT: Ray-Conditioned Visual Representations for Viewpoint-Robust Imitation Learning}

\author{
  \textbf{Qian Wang}\textsuperscript{1}\hspace{0.15em}\thanks{Email: \texttt{uxquy@student.kit.edu}}\quad
  \textbf{Longrui Chen}\textsuperscript{2} \quad
  \textbf{Peiran Sun}\textsuperscript{1} \quad
  \textbf{Aleksandar Taranovic}\textsuperscript{1} \\
  \textbf{Niklas Freymuth}\textsuperscript{1} \quad
  \textbf{Ge Li}\textsuperscript{1} \quad
  \textbf{Weiran Liao}\textsuperscript{1} \quad
  \textbf{C. F. Maximilian Nagy}\textsuperscript{1} \\
  \textbf{Yucheng Tang}\textsuperscript{1} \quad
  \textbf{Tao Chen}\textsuperscript{1} \quad
  \textbf{Gerhard Neumann}\textsuperscript{1} \\[0.6em]
  {\normalfont\small \textsuperscript{1}Karlsruhe Institute of Technology, Germany}\\[-0.1em]
  {\normalfont\small \textsuperscript{2}University of Leeds, United Kingdom}\\[-0.1em]
}

\begin{document}
\maketitle


\begin{abstract}
Visual imitation learning enables robots to acquire visuomotor skills directly from images, yet RGB observations lack explicit geometric cues, making learned policies brittle to camera perturbations. 
To address this, we propose \textbf{Ray-conditioned Vision Transformer Encoder (RayViT)}, a lightweight architecture that injects camera geometry into pretrained ViT backbones.
RayViT represents camera geometry as a Pl\"ucker ray map, patchifies it into ray features, and uses gated cross-attention to produce a ray-conditioned class token. 
These ray features are added as dense positional embeddings, while the ray class token replaces the original ViT class token to provide a geometry-aware summary representation. 
We combine this approach with an auxiliary cosine similarity loss to consistently improve the performance and robustness for geometry-aware tokens. Experiments on sim- and real-robot tasks demonstrate that RayViT improves robustness by approximately 13 percentage points under camera perturbations in multi-task RoboCasa benchmark and by 1.78 average completed stages in real-world multi-task success rate compared to baselines.

\end{abstract}



\section{Introduction}
\label{sec:introduction}

    Imitation learning enables robot policies to be trained from demonstrations of desired behaviors, usually recorded by human experts.
    These recordings come in different modalities, with RGB images being the most popular choice due to their semantic richness and low acquisition cost. 
    Benefiting from rich pretrained visual priors, powerful vision foundation models such as DINOv3~\citep{simeoni2025dinov3} consume these images and extract semantically meaningful information from demonstrations, making them a natural choice of visual encoder for downstream policies. 
    However, the resulting RGB-based policies are often brittle to camera movements and perturbations.
 
    Such movements arise because the camera must be moved when an obstacle obstructs its view, or because it is mounted to, e.g., a moving robot. 
    Changes in the camera parameters can significantly degrade policy performance at test time, since pretrained visual representations encode scene appearance but remain agnostic to the geometric relationship between camera and world. Consequently, policies built on these embeddings tend to learn viewpoint-specific appearance-to-action mappings rather than view-invariant task representations.
 
    Robustness to camera perturbations requires visual representations that are both camera-aware and consistent across viewpoints.
    Existing approaches rely on data diversity, collecting demonstrations from many cameras or viewpoints~\citep{pang2025reviwo,chen2024roviaug,tian2024vista,jiang2025knowcameraisviewinvariant}.
    While effective, they are expensive and hard to scale in real-world settings. 
    Alternative methods introduce explicit 3D representations such as point clouds or point maps to reduce viewpoint ambiguity~\citep{goyal2023rvt,goyal2024rvt2,ke2024diffuseractor,shridhar2023perceiver,jia2025pointmappolicy}, which provides useful geometric grounding but require additional sensors or reconstruction.
    Here, the density, noise and physical scale of 3D points are difficult to be unified.
    More critically, such 3D representations are not easily compatible with modern vision foundation models with pure-image priors.
    
    In this work, we instead draw on recent progress in novel view synthesis and dynamic scene reconstruction~\citep{wang2025uplvsm,szymanowicz2026lagernvs,wang2026raymap3r}, where camera intrinsics and extrinsics are encoded as Pl\"ucker ray embeddings~\citep{chengzhijing2026moca}. This encoding yields a \emph{ray map} in which each pixel is associated with its unique 3D viewing ray, represented by the ray's direction and spatial moment. Since the ray map is spatially aligned with the RGB image, it establishes a natural pixel-to-pixel correspondence between appearance and geometry. 
    Crucially, ray maps are computed from only intrinsic and extrinsic camera parameters and require no additional sensors, making them a lightweight geometric cue.
    The intrinsics are typically readily available through camera internal APIs, while camera extrinsics require world-frame calibration. The overhead for this calibration is aided by many readily available tools, and calibration is already required by many competing methods.

\begin{figure}[t!]    
    \centering
    \includegraphics[width=1\linewidth]{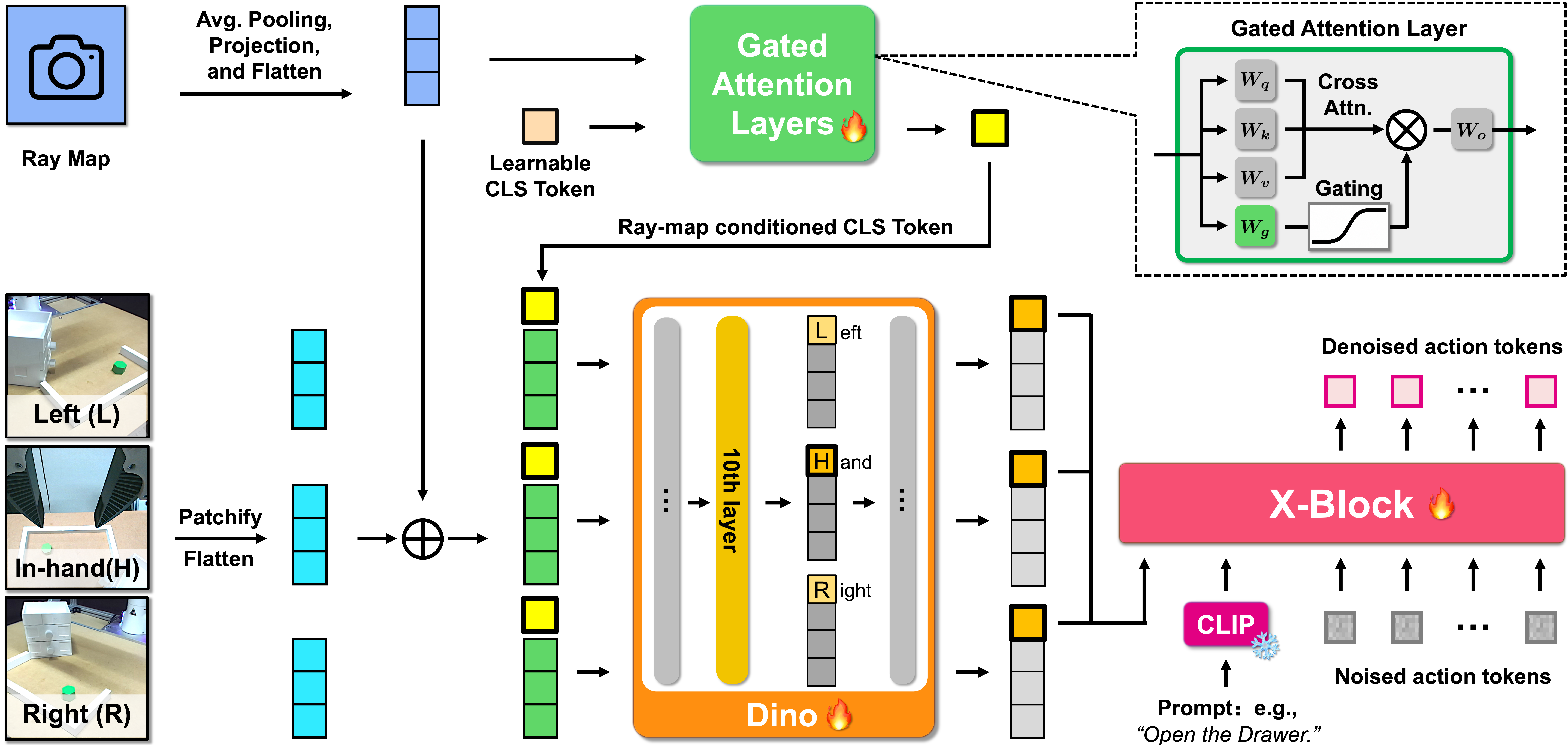}
    \caption{\textbf{Overview of RayViT}. Camera geometry is injected through a ray-map–conditioned class token and patch-aligned ray embeddings. An intermediate cross-view consistency loss regularizes the visual representation. Visual tokens, frozen CLIP \cite{radford2021learning} language tokens, and noisy action tokens are processed by a decoder-only xLSTM policy \cite{beck2024xlstmextendedlongshortterm, jia2025xilexploringdesignspace}.}
    \label{fig:architecture}
\end{figure}

    Having obtained a ray map, we find that injecting it into a pretrained ViT backbone is non-trivial. Naively concatenating geometric features with image patch tokens risks disrupting the pretrained representation distribution and degrading the rich semantic features acquired during pretraining. We therefore propose the Ray-conditioned Vision Transformer Encoder (RayViT), which injects camera geometry into a fully fine-tuned ViT encoder while preserving its pretrained structure. RayViT introduces two complementary geometry injection mechanisms. First, we propose a ray-map-conditioned class token. Instead of directly using DINOv3’s original CLS token, we condition a learnable token on the ray map through two gated cross-attention blocks\citep{qiu2025gatedattention}, producing a geometry-aware summary token while minimally perturbing the pretrained representation distribution. Figure~\ref{fig:architecture} provides a schematic overview. Second, we inject patch-aligned ray features as additional positional embeddings, providing explicit absolute 3D grounding for each patch token. Beyond camera awareness, we apply an auxiliary cosine similarity loss to geometry-conditioned class tokens at an intermediate layer, encouraging representations of the same scene from different viewpoints to converge. 
    This objective consistently improves both performance under the default setting and robustness to camera perturbations, but only when the tokens already carry geometric information.
    
    We evaluate RayViT in simulation and the real world. Across 16 simulated and 4 real-world manipulation tasks, our approach achieves competitive performance when considering fixed cameras. When the cameras are moved and thus change their parameters, we find that baseline methods drastically degrade, while RayViT exhibits only a minor performance drop and converges to significantly better performance than the baselines.
    
    In summary, our contributions are as follows: 
    1) We demonstrate that injecting camera geometry into pretrained ViT backbones substantially improves robustness to camera perturbations in imitation learning, without multi-view augmentation or additional sensors. 
    2) We propose RayViT, a lightweight architecture with a gated cross-attention module that injects Pl"ucker ray embeddings through a ray-conditioned class token and patch-level geometric positional embeddings.
    3) We show that view-consistency objective further yields consistent gains in both nominal performance and camera robustness, but only when applied to geometry-aware token representations.



	
\section{Related Work}
\label{sec:related_works}

\textbf{View-Invariant Representations from RGB.}
Pre-trained vision encoders have become standard backbones for imitation learning policies~\citep{nair2023r3m, xiao2022masked, majumdar2023we}, inheriting the broad capabilities of general-purpose vision foundation models~\citep{radford2021learning, kirillov2023segment, simeoni2025dinov3}.
Recent work pretrains backbones for 3D understanding~\citep{wang2025vggt}, and integrates them into manipulation policies~\citep{abouzeid2025geoawarevla,vuong2025evggt}.
However, neither family grounds its representations in the camera's pose relative to the workspace. 
The resulting policies tend to be brittle under visual distribution shifts and camera viewpoint changes~\citep{zhu2023learning, fei2025libero,jiang2025knowcameraisviewinvariant}.
While some methods are explicitly designed to be invariant to the viewpoint~\citep{simeonov2023se, zhang2025canonical,zhu2025equact}, they typically rely on point-cloud inputs and are thus not easily compatible with such pre-trained vision models.
Another line of work instead considers data augmentation via domain randomization~\citep{tobin2017domain}, image augmentation~\citep{laskin2020reinforcement,yarats2020image}, additional simulated cameras~\citep{pang2025reviwo} and novel views synthesized from existing demonstrations~\citep{chen2024roviaug,tian2024vista}.
Closely related to our work, \citet{jiang2025knowcameraisviewinvariant} also conditions the policy on Pl\"ucker ray maps, but still relies on instantiating many such cameras as a form of dataset augmentation.
Such augmentation is costly to scale, particularly outside simulation.
A different approach is taken by CLASS~\citep{lee2025class}, which in some settings operates on a fixed camera set like ours but follows a two-stage recipe in which an encoder is first pretrained with action-sequence contrastive learning and then frozen for policy training.
RayViT instead injects camera geometry at the representation level in a single end-to-end training, using only the cameras already present in the workspace. 
Our explicit ray-map conditioning and a cosine objective on the resulting tokens yield geometry-aware representations that remain robust under viewpoint changes.

\textbf{3D Representations and Viewpoint Robustness.}
A complementary line of work introduces explicit 3D structure into the
policy. The RVT family~\citep{goyal2023rvt,goyal2024rvt2} reconstructs
point clouds and re-renders the scene from fixed virtual cameras around
the workspace.
Several other methods skip re-rendering and feed 3D observations directly to the policy, including
raw point clouds~\citep{ze2024dp3, gervet2023act3d}, lifted RGB-D
features~\citep{ke2024diffuseractor, shridhar2023perceiver}, dense point
maps~\citep{jia2025pointmappolicy}, and Fourier-projected point
coordinates~\citep{gyenes2026fourier}. While these representations
encode scene geometry, the resulting policies still depend on the
extrinsics of the depth cameras used to construct the point cloud, and
generalize only modestly to novel camera viewpoints~\citep{wilcox2025adapt3r}.
Adapt3R~\citep{wilcox2025adapt3r} mitigates this issue by localizing 2D
backbone features in 3D relative to the end-effector, but requires
calibrated depth as an additional modality.
RayViT instead represents per-pixel camera geometry,
providing the pose-to-pixel correspondence these methods lack while
requiring no depth sensor and no point-cloud reconstruction.

\textbf{Ray Maps as a Geometric Cue.}
Pl\"{u}cker ray maps~\citep{chengzhijing2026moca, sitzmann2021light, jia2020plucker} map each pixel to a six-dimensional vector encoding
its ray's origin and direction in world coordinates, providing a dense
pose-to-pixel correspondence directly from camera intrinsics and
extrinsics. They have recently emerged as a compact way to encode camera
geometry across computer vision, including depth
estimation~\citep{lin2025depthanything3}, novel view
synthesis~\citep{wang2025uplvsm,szymanowicz2026lagernvs}, dynamic 3D
reconstruction~\citep{wang2026raymap3r}, and world
modeling~\citep{aether2025}. In imitation learning, ray-map conditioning
has so far been used only by \citet{jiang2025knowcameraisviewinvariant},
who use multi-view data augmentation and concatenate Pl\"{u}cker rays to image inputs or latent features.
This concatenation forces the encoder
to learn the patch-to-ray alignment from scratch.
RayViT instead injects ray geometry into the pretrained ViT token stream itself, with
patch-aligned ray features providing dense geometric positional context
and a gated cross-attention module producing a ray-conditioned class
token.

\section{Method}
\label{sec:method}

\textbf{Problem Formulation.}
Imitation learning aims to learn a policy from a dataset of expert demonstrations \mbox{\(D_{\tau} = \{\tau_i\}_{i=1}^{N}\)}. Each demonstration is represented as a sequence of observation-action pairs, \mbox{\(\tau_i = \{(o_1, a_1), (o_2, a_2), \ldots, (o_T, a_T)\}\), where \(o_t\)} and \(a_t\) denote the observation and expert action at time step \(t\), respectively. Besides conventional observations such as images \(I_{t}\), we incorporate ray maps \(r_{t}\) as additional observations, i.e., \mbox{\(I_{t}, r_{t} \in o_{t}\)} and learn a ray map conditioned visual representation \mbox{\(v_{t}= E(I_{t} | r_{t})\)}, where \(E\) denotes the encoder network. Following common visuomotor policy learning settings, the policy \(\pi\) is trained to predict an action or an action chunk \(a_{t:t+H}\) conditioned on the current observation \(o_t\). 


\textbf{Ray Map Modelling.}
We model camera geometry as ray map using Pl\"ucker coordinate representation, associating each image pixel with its corresponding 3D viewing ray determined by the camera intrinsics and extrinsics~\citep{chengzhijing2026moca, sitzmann2021light, jia2020plucker}. Let \mbox{$K \in \mathrm{R}^{3 \times 3}$} denote the camera intrinsic matrix, and let $(R, \mathbf{t})$ denote the world-to-camera extrinsic, where $R \in \mathrm{SO}(3)$ and $\mathbf{t} \in \mathrm{R}^{3}$. For a pixel $\mathbf{p} = (u, v, 1)^{\top}$, the camera center and viewing direction in the world frame are computed as
\begin{equation}
    \mathbf{o}_{w} = -R^{\top}\mathbf{t},
    \qquad
    \tilde{\mathbf{d}}_{w}(u,v) = R^{\top}K^{-1}\mathbf{p},
    \qquad
    \mathbf{d}_{w}(u,v) = \frac{\tilde{\mathbf{d}}_{w}(u,v)}{\|\tilde{\mathbf{d}}_{w}(u,v)\|_2}.
\end{equation}
Following the Pl\"ucker line representation, each ray is encoded by its direction and moment,
\begin{equation}
    \mathbf{m}_{w}(u,v) = \mathbf{o}_{w} \times \mathbf{d}_{w}(u,v),
    \qquad
    \mathbf{r}(u,v) = [\mathbf{d}_{w}(u,v), \mathbf{m}_{w}(u,v)] \in \mathrm{R}^{6}.
\end{equation}
Stacking $\mathbf{r}(u,v)$ over all pixels yields a ray map spatially aligned with the input image, providing a dense per-pixel geometric representation of the camera viewpoint.


\textbf{Ray Map conditioned Class Token.}
In standard ViTs, the class token [CLS] is a learnable token prepended to the patch sequence to aggregate global visual information through self-attention. The resulting [CLS] is typically passed to the policy network as a global visual descriptor for action prediction. Since [CLS] is initialized as a free parameter, its representation is built solely from interactions with image patches. Under camera shifts, small pose changes can significantly alter image appearance and spatial layout, while [CLS] lacks direct access to the underlying camera geometry that explains these changes. To address this, we first propose the \textit{ray-map-conditioned class token}.


\textbf{Patch-aligned Ray Features}. We downsample the ray map to the encoder patch resolution using average pooling with kernel size equal to the patch size. Unlike learned convolutions, average pooling is non-parametric and better preserves the geometric semantics of Plücker coordinates. The resulting ray grid is spatially aligned with the encoder patch tokens in a one-to-one manner, providing a natural mechanism for injecting geometry-aware positional embeddings into the image patch tokens.
We then project the 6-dimensional ray-patch features to the encoder's latent dimension through a two-layer MLP with GELU activation.

\textbf{Gated Cross-attention Module.}
The ray-map initializer consists of two lightweight gated cross-attention blocks matching the hidden dimension of the pretrained ViT backbone. A learnable query token cross-attends to the patch-aligned ray features, progressively summarizing camera geometry into a single token. Following Qwen~\citep{qiu2025gatedattention}, we apply a learnable element-wise sigmoid gate after attention to regulate geometric conditioning. This control is important because the ray map should adaptively condition the global token without overwhelming the pretrained representational signal that subsequent transformer blocks expect. The final query token is used as the ray-conditioned class token, replacing the original [CLS] token in the ViT forward pass.



\textbf{Ray Map Positional Embedding.}
\label{subsec:raype}
While RGB images capture scene appearance, ray maps encode the camera geometry underlying each observation. To build a camera-aware visual representation, we also inject ray features into the ViT patch-token stream.

\textbf{Additive injection alongside RoPE.} We add the patch-aligned ray features to the image patch tokens at the encoder input, before the first transformer block. We retain the original rotary positional embedding (RoPE)~\citep{simeoni2025dinov3} in backbone unchanged, as it is complementary to our design. RoPE supports local spatial reasoning in the image plane, whereas our ray-based signal provides absolute grounding for each patch. Because RoPE is camera-agnostic, it cannot capture viewpoint-dependent 3D structure under varying camera poses. Our additive ray injection fills this gap while preserving the pretrained attention dynamics, since RoPE remains untouched inside the attention mechanism.
  

\textbf{Cross-view Alignment Objective.}
To encourage viewpoint-robust representations, we align the class tokens across views such that they capture shared manipulation-relevant content despite differing camera geometries. 
We use the in-hand class token as the alignment anchor, since the gripper view is closest to the manipulation target and contains less irrelevant background than third-person views. Concretely, we apply a cosine-similarity loss to the class tokens extracted from an intermediate layer of the visual backbone. For \(N\) camera views, let \(g\) denote a reference view, and let \(\mathcal{V}=\{v_1,\ldots,v_{N-1}\}\) denote the remaining non-reference views. We use \(g\) as the anchor and define
\begin{equation}
\mathcal{L}_{\mathrm{cos}} = 1 - \frac{1}{N-1}\sum_{v \in \mathcal{V}} \cos\left(c^{v}, c^{g}\right),
\qquad
\mathcal{L} = \mathcal{L}_{\mathrm{IL}} + \lambda \, \mathcal{L}_{\mathrm{cos}}.
\end{equation}
Here, \(\cos(\cdot, \cdot)\) denotes cosine similarity, \(c^{g}\) denotes the reference class token, \(c^{v}\) denotes the class token of another camera view, and \(\lambda\) is a scalar weight, default value in Table~\ref{tab:rayvit_hyperparameters}. The cosine alignment term is used only as an auxiliary regularizer and is added to the primary policy learning loss.




\section{Simulated Experiments}
\label{sec:experiments}
	Following a multi-task setting, we conduct simulation experiments on the RoboCasa benchmark using 16 tasks that cover diverse manipulation skills.

	\textbf{Robocasa\citep{robocasa2024}}. It is a large-scale benchmark for robotic manipulation in realistic kitchen environments. It provides diverse scenes, layouts, object assets, and manipulation tasks ranging from simple actions to long-horizon, multi-step behaviors. Given large-scale demonstrations, RoboCasa serves as a challenging testbed for developing and evaluating imitation learning policies.

	\textbf{Experimental Setup}. To ensure a fair comparison across different visual representations, all methods use the same score-based diffusion policy with a decoder-only xLSTM action head. Experiments are conducted in a multi-task setting, where a single policy is trained and evaluated across 16 tasks. Unless otherwise specified, the visual encoder is fully fine-tuned end-to-end jointly with the policy. Each model is trained for 50 epochs with 3 random seeds, and we evaluate checkpoints at 30th, 40th, and 50th epoch, reporting the best result. 

	To evaluate robustness under camera perturbations, we test methods both on the training camera setup and the perturbed camera configurations. We construct a variant simulation environment in which the two static cameras are randomly rotated within bounded azimuth and elevation ranges, corresponding to horizontal and vertical viewing-angle changes, and also randomly shifted in distance relative to the reference fixture.
	
	\textbf{Baselines and Variants}. We provide two variants of our method with different levels of camera geometry injection. \textit{RayViT-cls} injects camera geometry only through a ray-map–conditioned class token, whereas \textit{RayViT} further incorporates patch-aligned ray features as positional embeddings. We always use the gripper camera view as the anchor $g$ for the cross-view alignment objective.
	We compare both \textit{RayViT} variants against four representative baselines, namely Adapt3R~\citep{wilcox2025adapt3r}, PMP-xyz~\citep{jia2025pointmappolicy}, CamPose~\citep{jiang2025knowcameraisviewinvariant} and our image-only variant without any ray-map conditioning. 
	Detailed descriptions of the baselines are provided in Appendix~\ref{subsec: baseline introduction}.


	\textbf{Results}
\begin{table*}[t!]
\begin{center}
\resizebox{\textwidth}{!}{%
\renewcommand{\arraystretch}{1.3}
\setlength{\aboverulesep}{0pt}
\setlength{\belowrulesep}{0pt}
\begin{tabular}{l|cc|cc|cc|cc|cc|cc}
\toprule
\multirow{2}{*}{\textbf{Task}} & \multicolumn{2}{c|}{Adapt3R} & \multicolumn{2}{c|}{PMP-xyz} & \multicolumn{2}{c|}{CamPose} & \multicolumn{2}{c|}{RGB} & \multicolumn{2}{c|}{\textbf{RayVit-cls}} & \multicolumn{2}{c}{\textbf{RayVit}}\\
\cmidrule(lr){2-3}\cmidrule(lr){4-5}\cmidrule(lr){6-7}\cmidrule(lr){8-9}\cmidrule(lr){10-11}\cmidrule(lr){12-13}

& Default & Variant & Default & Variant & Default & Variant & Default & Variant & Default & Variant & Default & Variant \\
\midrule

PnPCounterToMicrowave
& \bestgroupone{$9.3 \scriptstyle \pm 0.9$}
& $0.7 \scriptstyle \pm 0.9$
& $2.0 \scriptstyle \pm 0.0$
& $0.0 \scriptstyle \pm 0.0$
& $0.0 \scriptstyle \pm 0.0$
& $0.0 \scriptstyle \pm 0.0$
& $2.0 \scriptstyle \pm 1.6$
& $0.0 \scriptstyle \pm 0.0$
& $1.3 \scriptstyle \pm 0.9$
& $0.0 \scriptstyle \pm 0.0$
& $2.0 \scriptstyle \pm 1.6$
& \bestgrouptwo{$2.0 \scriptstyle \pm 1.6$}
\\
PnPCounterToSink
& \bestgroupone{$2.7 \scriptstyle \pm 1.9$}
& $0.0 \scriptstyle \pm 0.0$
& $2.0 \scriptstyle \pm 2.0$
& \bestgrouptwo{$2.0 \scriptstyle \pm 0.0$}
& $0.0 \scriptstyle \pm 0.0$
& $0.0 \scriptstyle \pm 0.0$
& $0.7 \scriptstyle \pm 0.9$
& $0.7 \scriptstyle \pm 0.9$
& $0.7 \scriptstyle \pm 0.9$
& $1.3 \scriptstyle \pm 1.9$
& \bestgroupone{$2.7 \scriptstyle \pm 0.9$}
& $1.3 \scriptstyle \pm 1.9$
\\
PnPMicrowaveToCounter
& $6.0 \scriptstyle \pm 1.6$
& $3.3 \scriptstyle \pm 1.9$
& $10.0 \scriptstyle \pm 2.0$
& $2.7 \scriptstyle \pm 3.1$
& $1.3 \scriptstyle \pm 1.2$
& $0.7 \scriptstyle \pm 1.2$
& \bestgroupone{$14.0 \scriptstyle \pm 1.6$}
& $0.7 \scriptstyle \pm 0.9$
& $13.3 \scriptstyle \pm 0.9$
& \bestgrouptwo{$8.7 \scriptstyle \pm 3.4$}
& $6.7 \scriptstyle \pm 2.5$
& $6.7 \scriptstyle \pm 3.4$
\\
PnPSinkToCounter
& $3.3 \scriptstyle \pm 0.9$
& $4.7 \scriptstyle \pm 0.9$
& $4.7 \scriptstyle \pm 1.2$
& $2.0 \scriptstyle \pm 2.0$
& $3.3 \scriptstyle \pm 1.2$
& $2.0 \scriptstyle \pm 2.0$
& \bestgroupone{$18.0 \scriptstyle \pm 1.6$}
& $6.0\scriptstyle \pm 2.8$
& $13.3 \scriptstyle \pm 5.2$
& \bestgrouptwo{$12.0 \scriptstyle \pm 4.3$}
& $9.3 \scriptstyle \pm 5.0$
& $6.0 \scriptstyle \pm 3.3$
\\
\midrule
OpenDrawer
& $47.3 \scriptstyle \pm 3.4$
& $30.0 \scriptstyle \pm 7.5$
& \bestgroupone{$49.3 \scriptstyle \pm 1.2$}
& $18.0 \scriptstyle \pm 7.2$
& $31.3 \scriptstyle \pm 4.2$
& $35.3 \scriptstyle \pm 3.1$
& $44.7 \scriptstyle \pm 5.2$
& $14.0 \scriptstyle \pm 3.3$
& $44.7 \scriptstyle \pm 6.2$
& \bestgrouptwo{$40.7 \scriptstyle \pm 5.0$}
& $34.7 \scriptstyle \pm 6.8$
& $29.3 \scriptstyle \pm 5.0$
\\
CloseDrawer
& $82.0 \scriptstyle \pm 1.6$
& $67.3 \scriptstyle \pm 9.3$
& \bestgroupone{$99.3 \scriptstyle \pm 1.2$}
& $55.3 \scriptstyle \pm 23.0$
& $82.7 \scriptstyle \pm 5.8$
& \bestgrouptwo{$84.7 \scriptstyle \pm 5.8$}
& $86.7 \scriptstyle \pm 1.9$
& $42.0 \scriptstyle \pm 7.1$
& $87.3 \scriptstyle \pm 2.5$
& $81.3 \scriptstyle \pm 4.1$
& $81.3 \scriptstyle \pm 9.4$
& \bestgrouptwo{$84.7 \scriptstyle \pm 4.1$}
\\
\midrule

TurnOnStove
& \bestgroupone{$37.3 \scriptstyle \pm 1.9$}
& $20.7 \scriptstyle \pm 7.4$
& $34.7 \scriptstyle \pm 6.4$
& $8.7 \scriptstyle \pm 1.2$
& $17.3 \scriptstyle \pm 3.1$
& $13.3 \scriptstyle \pm 1.2$
& $24.7 \scriptstyle \pm 2.5$
& $14.0 \scriptstyle \pm 4.3$
& $28.0 \scriptstyle \pm 0.0$
& $24.0 \scriptstyle \pm 5.9$
& $28.7 \scriptstyle \pm 1.9$
& \bestgrouptwo{$30.7 \scriptstyle \pm 2.5$}
\\
TurnOffStove
& \bestgroupone{$18.0 \scriptstyle \pm 1.6$}
& $12.0 \scriptstyle \pm 4.3$
& $16.0 \scriptstyle \pm 2.0$
& $7.3 \scriptstyle \pm 9.2$
& $8.7 \scriptstyle \pm 2.3$
& $10.0 \scriptstyle \pm 0$
& $11.3 \scriptstyle \pm 1.9$
& $9.3 \scriptstyle \pm 2.5$
& $14.7 \scriptstyle \pm 2.5$
& $13.3 \scriptstyle \pm 2.5$
& $14.0 \scriptstyle \pm 1.6$
& \bestgrouptwo{$17.3 \scriptstyle \pm 0.9$}
\\
\midrule

TurnOnSinkFaucet
& $68.0 \scriptstyle \pm 4.3$
& $46.0 \scriptstyle \pm 9.9$
& $68.7 \scriptstyle \pm 2.3$
& $33.3 \scriptstyle \pm 10.1$
& $22.0 \scriptstyle \pm 2.0$
& $20.0 \scriptstyle \pm 5.3$
& $52.7 \scriptstyle \pm 7.7$
& $42.7\scriptstyle \pm 9.0$
& \bestgroupone{$70.0 \scriptstyle \pm 3.3$}
& \bestgrouptwo{$68.0 \scriptstyle \pm 9.9$}
& $65.3 \scriptstyle \pm 2.5$
& $66.0 \scriptstyle \pm 4.9$
\\
TurnOffSinkFaucet
& $69.3 \scriptstyle \pm 5.0$
& $42.0 \scriptstyle \pm 11.8$
& \bestgroupone{$76.0 \scriptstyle \pm 6.9$}
& $36.7 \scriptstyle \pm 1.2$
& $36.0 \scriptstyle \pm 7.2$
& $34.7 \scriptstyle \pm 5.0$
& $60.7 \scriptstyle \pm 6.8$
& $30.0 \scriptstyle \pm 4.3$
& $52.7 \scriptstyle \pm 5.2$
& $48.7 \scriptstyle \pm 5.2$
& $56.7 \scriptstyle \pm 4.1$
& \bestgrouptwo{$51.3 \scriptstyle \pm 9.0$}
\\
TurnSinkSpout
& $65.3 \scriptstyle \pm 0.9$
& \bestgrouptwo{$52.7 \scriptstyle \pm 4.1$}
& \bestgroupone{$72.0 \scriptstyle \pm 2.0$}
& $48.0 \scriptstyle \pm 6.9$
& $10.7 \scriptstyle \pm 1.2$
& $7.3 \scriptstyle \pm 1.2$
& $40.7 \scriptstyle \pm 3.4$
& $40.7 \scriptstyle \pm 4.1$
& $40.7 \scriptstyle \pm 7.4$
& $43.3 \scriptstyle \pm 3.4$
& $42.7 \scriptstyle \pm 3.8$
& $43.3 \scriptstyle \pm 10.5$
\\
\midrule

CoffeePressButton
& $73.3 \scriptstyle \pm 3.8$
& $58.0 \scriptstyle \pm 5.7$
& $75.3 \scriptstyle \pm 8.1$
& $53.3 \scriptstyle \pm 8.1$
& $46.7 \scriptstyle \pm 1.2$
& $40.7 \scriptstyle \pm 1.2$
& $90.7 \scriptstyle \pm 1.9$
& $80.7 \scriptstyle \pm 4.1$
& \bestgroupone{$92.0 \scriptstyle \pm 4.3$}
& \bestgrouptwo{$85.3 \scriptstyle \pm 5.0$}
& $84.7 \scriptstyle \pm 4.7$
& $84.7 \scriptstyle \pm 2.5$
\\
TurnOnMicrowave
& $64.7 \scriptstyle \pm 1.9$
& $40.0 \scriptstyle \pm 7.1$
& $46.0 \scriptstyle \pm 8.7$
& $10.7 \scriptstyle \pm 5.0$
& $29.3 \scriptstyle \pm 1.2$
& $26.7 \scriptstyle \pm 10.3$
& \bestgroupone{$80.0 \scriptstyle \pm 0.0$}
& $40.0 \scriptstyle \pm 10.2$
& $72.0 \scriptstyle \pm 5.7$
& $48.0 \scriptstyle \pm 7.1$
& $64.0 \scriptstyle \pm 4.3$
& \bestgrouptwo{$54.0 \scriptstyle \pm 2.8$}
\\
TurnOffMicrowave
& $69.3 \scriptstyle \pm 4.7$
& $44.7 \scriptstyle \pm 2.5$
& $50.7 \scriptstyle \pm 9.0$
& $7.3 \scriptstyle \pm 4.6$
& $32.0 \scriptstyle \pm 7.2$
& $36.7 \scriptstyle \pm 2.3$
& $88.7 \scriptstyle \pm 5.0$
& $40.7 \scriptstyle \pm 9.8$
& \bestgroupone{$91.3 \scriptstyle \pm 2.5$}
& \bestgrouptwo{$62.7 \scriptstyle \pm 12.0$}
& $85.3 \scriptstyle \pm 2.5$
& $61.3 \scriptstyle \pm 10.9$
\\
\midrule

CoffeeServeMug
& $38.7 \scriptstyle \pm 2.5$
& $26.0 \scriptstyle \pm 8.5$
& \bestgroupone{$52.7 \scriptstyle \pm 8.1$}
& $33.3 \scriptstyle \pm 6.4$
& $16.7 \scriptstyle \pm 3.1$
& $18.0 \scriptstyle \pm 5.3$
& $51.3 \scriptstyle \pm 3.4$
& $16.7 \scriptstyle \pm 5.0$
& $52.0 \scriptstyle \pm 3.3$
& $42.0 \scriptstyle \pm 9.9$
& $50.7 \scriptstyle \pm 2.5$
& \bestgrouptwo{$46.7 \scriptstyle \pm 4.7$}
\\
CoffeeSetupMug
& $12.7 \scriptstyle \pm 0.9$
& $8.0 \scriptstyle \pm 3.3$
& $14.7 \scriptstyle \pm 1.2$
& \bestgrouptwo{$14.0 \scriptstyle \pm 2.0$}
& $10.0 \scriptstyle \pm 5.3$
& $12.7 \scriptstyle \pm 3.1$
& \bestgroupone{$16.7 \scriptstyle \pm 4.7$}
& $6.0 \scriptstyle \pm 3.3$
& \bestgroupone{$16.7 \scriptstyle \pm 4.7$}
& $12.7 \scriptstyle \pm 1.9$
& $7.3 \scriptstyle \pm 3.4$
& $10.7 \scriptstyle \pm 1.9$
\\
\midrule
{\textbf{Average Success Rate $\boldsymbol{\uparrow}$}}
& $41.7$
& $28.5 $
& $42.1 $
& $20.8 $
& $21.8 $
& $21.4 $
& $42.5$
& $24.0$
& \bestgroupone{$43.2$}
& $37.0$
& $39.8$
& \bestgrouptwo{$37.3$}
\\
\midrule

{\textbf{Performance Drop} $\boldsymbol{\downarrow}$}
& $-$
& $13.2$
& $-$
& $21.3 $
& $-$
& $0.4 $
& $-$
& $18.5$
& $-$
& $6.2$
& $-$
& $2.5$
\\
\bottomrule
\end{tabular}
}
\end{center}
\caption{Success rate (\%) for each task in RoboCasa~\cite{robocasa2024}. The models were evaluated over 50 episodes per task. Best results in the \textbf{\textcolor{groupone}{Default}} and \textbf{\textcolor{grouptwo}{Variant}} settings are highlighted in \textbf{\textcolor{groupone}{blue}} and \textbf{\textcolor{grouptwo}{green}}, respectively. $\boldsymbol{\uparrow}$ indicates that a higher average success rate is better, while $\boldsymbol{\downarrow}$ indicates that a lower performance drop between the default and variant settings is better.}
\label{table:main_result}
\end{table*}
	We present the empirical results in Table~\ref{table:main_result}, comparing all methods under both default and variant camera settings. Under the default setting, RayViT-cls and RayViT achieve performance comparable to the strongest baselines, including the RGB-only variant, with RayViT-cls even slightly outperforming all baselines. These results indicate that our geometry-conditioned visual representation effectively benefits policy learning, despite minor modifications to the pretrained visual backbone, which may disrupt the pretrained feature distribution.
	
	Under random perturbations, our methods outperform the 2D RGB baseline by at least 13 percentage points in average success rate and consistently surpass explicit 3D representations, indicating that dense geometry alone is insufficient for viewpoint robustness without explicitly modeling camera-induced transformations.
	Compared with Adapt3R, direct ray-map conditioning provides a stronger inductive bias without requiring additional sensing modalities. Although CamPose shows only a modest performance drop under perturbation, its lower performance in both settings suggests that late-fused ray-map conditioning is less effective than directly integrating camera geometry into the visual representation, particularly in the multi-task setting without multi-view data augmentation.


	\subsection{Parameter Study and Analysis}
\begin{wraptable}{r}{0.40\textwidth}
\vspace{-1.0em}
\centering
\small
\resizebox{\linewidth}{!}{%
\renewcommand{\arraystretch}{1.15}
\setlength{\aboverulesep}{0pt}
\setlength{\belowrulesep}{0pt}
\begin{tabular}{l|ccc}
\toprule
Method & Default & Variant & Drop $\boldsymbol{\downarrow}$ \\
\midrule

RGB             & $42.5 \scriptstyle \pm 0.8$ & $24.0\scriptstyle \pm 0.5$ & 18.5 \\

\midrule

RayViT-cls w/o. & $38.4 \scriptstyle \pm 1.7$ & $30.6 \scriptstyle \pm 8.7$ & 7.8 \\
RayViT-cls      & $43.2 \scriptstyle \pm 1.7$ & $37.0 \scriptstyle \pm 3.5$ & 6.2 \\

\midrule

RayViT w/o.     & $39.5 \scriptstyle \pm 4.4$ & $35.6 \scriptstyle \pm 9.1$ & 3.9 \\
RayViT          & $39.8 \scriptstyle \pm 1.9$ & $37.3 \scriptstyle \pm 1.2$ & 2.5 \\

\bottomrule
\end{tabular}
}
\caption{Effect of ray-map conditioning. \textit{W/o.} denotes without the cosine similarity loss.}
\label{table:ablation_q1}
\vspace{-1.0em}
\end{wraptable}

	\textbf{Effect of Ray-Map Conditioning}. Table~\ref{table:ablation_q1} reports the average success rate across 16 tasks for models trained with and without the auxiliary cosine similarity loss. Compared with the RGB variant, geometry injection alone already improves robustness to camera perturbations. However, its slightly lower performance under the default setting suggests that ray-map conditioning may perturb the pretrained feature distribution and introduce a small optimization disadvantage.

	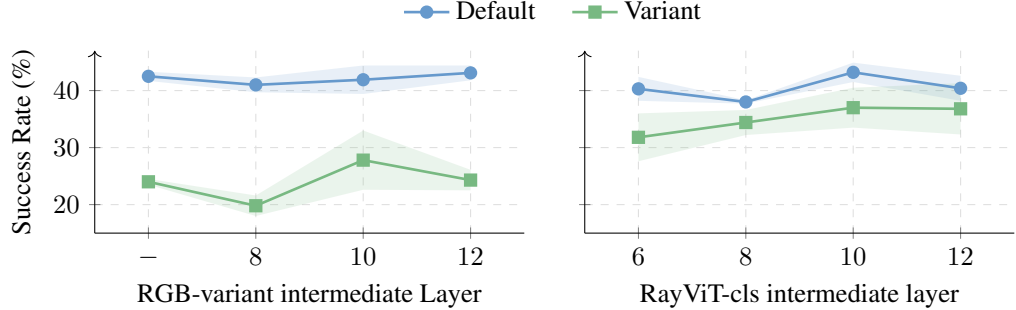
\begin{figure}[t]
    \centering
    \begin{tikzpicture}
        \begin{groupplot}[
            group style={
                group size=2 by 1,
                horizontal sep=0.8cm,
                ylabels at=edge left,
                yticklabels at=edge left,
            },
            width=0.52\linewidth,
            height=4.0cm,
            ylabel={Success Rate (\%)},
            ymin=15, ymax=47,
            ytick={20,30,40},
            xticklabel style={font=\normalsize, anchor=north},
            yticklabel style={font=\normalsize},
            grid=major,
            grid style={dashed, gray!25},
            tick align=outside,
            tick pos=left,
            axis lines=left,
            axis line style={->},
            x axis line style={-},
            label style={font=\normalsize},
            every axis plot/.append style={line width=1pt, mark size=2pt},
        ]
        \nextgroupplot[
            xlabel={RGB-variant intermediate Layer},
            xmin=5, xmax=13,
            xtick={6, 8, 10, 12},
            xticklabels={$-$, $8$, $10$, $12$},
            legend to name=sharedlegend,
            legend style={
                legend columns=2,
                font=\normalsize,
                draw=none,
                /tikz/every even column/.append style={column sep=0.4cm},
            },
        ]
        \addplot[draw=none, fill={rgb,255:red,102;green,153;blue,204}, fill opacity=0.15, forget plot]
            coordinates {
            (6, 43.3) (8, 42.3) (10, 44.4) (12, 44.4)
            (12, 41.8) (10, 39.4) (8, 39.7) (6, 41.7)
        } \closedcycle;
        \addplot[draw=none, fill={rgb,255:red,120;green,185;blue,130}, fill opacity=0.15, forget plot]
            coordinates {
            (6, 24.5) (8, 21.6) (10, 33.0) (12, 26.1)
            (12, 22.5) (10, 22.6) (8, 18.0) (6, 23.5)
        } \closedcycle;

        \addplot[color={rgb,255:red,102;green,153;blue,204}, mark=*] coordinates {
            (6, 42.5) (8, 41.0) (10, 41.9) (12, 43.1)
        };
        \addlegendentry{Default}
        \addplot[color={rgb,255:red,120;green,185;blue,130}, mark=square*] coordinates {
            (6, 24.0) (8, 19.8) (10, 27.8) (12, 24.3)
        };
        \addlegendentry{Variant}

        \nextgroupplot[
            xlabel={RayViT-cls intermediate layer},
            xmin=5, xmax=13,
            xtick={6, 8, 10, 12},
            xticklabels={$6$, $8$, $10$, $12$},
        ]
        \addplot[draw=none, fill={rgb,255:red,102;green,153;blue,204}, fill opacity=0.15, forget plot]
            coordinates {
            (6, 42.4) (8, 38.3) (10, 44.9) (12, 42.6)
            (12, 38.2) (10, 41.5) (8, 37.7) (6, 38.2)
        } \closedcycle;
        \addplot[draw=none, fill={rgb,255:red,120;green,185;blue,130}, fill opacity=0.15, forget plot]
            coordinates {
            (6, 36.0) (8, 36.6) (10, 40.5) (12, 41.3)
            (12, 32.3) (10, 33.5) (8, 32.2) (6, 27.6)
        } \closedcycle;

        \addplot[color={rgb,255:red,102;green,153;blue,204}, mark=*] coordinates {
            (6, 40.3) (8, 38.0) (10, 43.2) (12, 40.4)
        };
        \addplot[color={rgb,255:red,120;green,185;blue,130}, mark=square*] coordinates {
            (6, 31.8) (8, 34.4) (10, 37.0) (12, 36.8)
        };
        \end{groupplot}
        \node[anchor=south] at ($(group c1r1.north)!0.5!(group c2r1.north) + (0,0.1cm)$)
            {\pgfplotslegendfromname{sharedlegend}};
    \end{tikzpicture}
    \caption{RGB-variant and RayViT-cls success rates when applying the auxiliary cosine-similarity loss at different intermediate layers.
    Colored lines report mean success rate, with background bands indicating standard deviation.
    \textbf{Left:} Effect of auxiliary loss on the RGB variant. \texttt{-} denotes no auxiliary loss applied at an intermediate layer.
    \textbf{Right:} Influence of the auxiliary loss when injected into RayViT-cls layers $6{-}12$.}
    \label{fig:ablation_loss_layer}
\end{figure}
	\textbf{Role of the View-Consistency Objective}. The results in Table~\ref{table:ablation_q1} further show that RayViT-cls is more sensitive to the removal of the cross-view objective, while also benefiting more strongly from it than RayViT. This may be because RayViT already incorporates richer geometric information by injecting ray features as additional positional embeddings for image patches. Interestingly, the auxiliary objective consistently improves performance under both default and variant settings.
	However, as shown on the left of Figure~\ref{fig:ablation_loss_layer}, applying the same cosine similarity loss to the RGB-vanilla baseline does not yield stable performance gains, suggesting that the effectiveness of the cross-view objective depends on geometry-aware representations.

	\textbf{Layer-wise Application of the Auxiliary Loss}. We further study the effect of the layer at which the view-consistency constraint is applied to the global class tokens of the pretrained ViT backbone.
	As shown on the right side of Figure~\ref{fig:ablation_loss_layer}, applying the auxiliary loss at deeper layers leads to the strongest gains in both default performance and robustness under camera perturbations, whereas applying it at earlier layers yields only marginal improvements over the variant without the view-consistency objective.
	
\begin{wraptable}{r}{0.40\textwidth}
\centering
\small
\resizebox{\linewidth}{!}{%
\renewcommand{\arraystretch}{1.15}
\setlength{\aboverulesep}{0pt}
\setlength{\belowrulesep}{0pt}
\begin{tabular}{l|ccc}
\toprule
Method & Default & Variant & Drop $\boldsymbol{\downarrow}$\\
\midrule

RGB scr   & $29.2 \scriptstyle \pm 0.4$ & $10.0 \scriptstyle \pm 2.2$ & 19.2 \\
RayViT-cls scr& $33.2 \scriptstyle \pm 1.6$ & $28.2 \scriptstyle \pm 1.4$ & 5.0 \\
RayViT scr & $32.1 \scriptstyle \pm 1.0$ & $29.2 \scriptstyle \pm 0.9$ & 2.9 \\

\bottomrule
\end{tabular}
}
\caption{Ablation on training visual backbone without pretrained weights; scr denotes "trained from scratch"}
\label{table:ablation_q4}
\vspace{-1.0em}
\end{wraptable}
	
	\textbf{Without pretrained Weights}. Although our methods are designed on top of a pretrained ViT backbone, we further examine whether the advantages of our methods remain when training without pretrained weights. The Table~\ref{table:ablation_q4} shows that our geometry-aware representation continues to outperform pure image representations, indicating that its benefits are not solely dependent on pretrained distribution.

\begin{wraptable}{r}{0.40\textwidth}
\vspace{-1.0em}
\centering
\small
\resizebox{\linewidth}{!}{%
\renewcommand{\arraystretch}{1.15}
\setlength{\aboverulesep}{0pt}
\setlength{\belowrulesep}{0pt}
\begin{tabular}{l|ccc}
\toprule
Method & Default & Variant & Drop $\boldsymbol{\downarrow}$ \\
\midrule

RGB        & $44.1 \scriptstyle \pm 0.2$ & $18.9 \scriptstyle \pm 2.5$ & 25.2 \\
RayViT-cls & $44.2 \scriptstyle \pm 2.2$ & $26.6 \scriptstyle \pm 4.2$ & 17.6 \\
RayViT     & $44.3 \scriptstyle \pm 0.6$ & $33.3 \scriptstyle \pm 5.6$ & 11.0 \\

\bottomrule
\end{tabular}
}
\caption{Ablation on different pretrained ViT backbone(EUPE). The results are still average success rate from 16 tasks in RoboCasa with 3 seeds.}
\label{table:ablation_q5}
\vspace{-1.0em}
\end{wraptable}

	\textbf{Transferability across pretrained ViT Backbones}. Our experiments primarily use pretrained DINOv3 \cite{simeoni2025dinov3} as the visual backbone. To evaluate the generality of our approach, we further evaluate our methods on the Efficient Universal Perception Encoder (EUPE) \cite{zhu2026eupe}, a ViT-based model distilled from multiple domain-expert foundation vision models. As shown in Table~\ref{table:ablation_q5}, our methods remain effective across different pretrained backbones, improving performance under both default and variant settings. Interestingly, RayViT outperforms RayViT-cls with EUPE, unlike the DINOv3 setting, suggesting that richer geometric conditioning is more beneficial under diverse pretrained feature distributions.

\section{Real Robot Experiments}

\begin{figure*}[t!]
\centering
\begin{minipage}[t]{0.50\textwidth}
    \vspace{0pt}
    \centering
    \includegraphics[width=\linewidth, trim=0 90 0 0, clip]{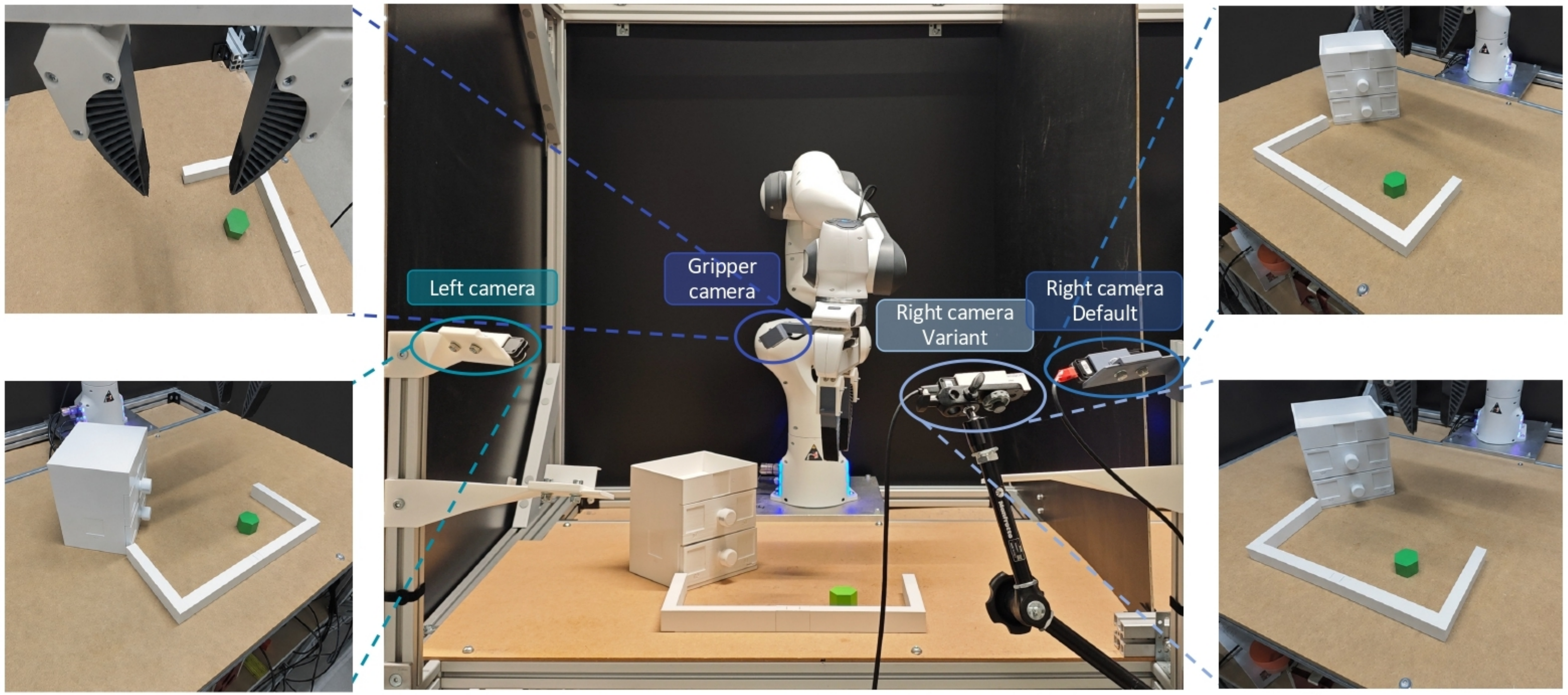}
    \vspace{-0.8em}
    \caption{Real-robot experimental setup. Vision system uses three cameras: left-, right-, and gripper-views and adapts a variant right-side view}
    \label{fig:real_robot_setup}
\end{minipage}
\hspace{0.005\textwidth}
\begin{minipage}[t]{0.48\textwidth}
    \vspace{0.8pt}
    \centering
    \resizebox{\linewidth}{!}{%
\renewcommand{\arraystretch}{1.3}
\setlength{\aboverulesep}{0pt}
\setlength{\belowrulesep}{0pt}
\begin{tabular}{l|c|cc|cc|cc}
\toprule
\multirow{2}{*}{\textbf{Task}} & \multirow{2}{*}{\textbf{Max}} & \multicolumn{2}{c|}{\textbf{RGB}} & \multicolumn{2}{c|}{\textbf{RayViT-cls}} & \multicolumn{2}{c}{\textbf{RayViT}}\\
\cmidrule(lr){3-4}\cmidrule(lr){5-6}\cmidrule(l){7-8}
& & Default & Variant & Default & Variant & Default & Variant\\
\midrule
Drawer     & $4$ & $2.95$ & $0.45$ & $1.80$ & $1.45$ & \bestgroupone{$3.45$} & \bestgrouptwo{$2.80$}\\
Insertion  & $3$ & $1.55$ & $1.25$ & \bestgroupone{$2.20$} & $0.60$ & $1.70$ & \bestgrouptwo{$1.50$}\\
Pick Place & $4$ & $1.85$ & $0.90$ & $1.60$ & $1.50$ & \bestgroupone{$2.60$} & \bestgrouptwo{$3.10$}\\
Cup Stack  & $4$ & $3.40$ & $1.05$ & $3.45$ & $3.25$ & \bestgroupone{$3.50$} & \bestgrouptwo{$3.35$}\\
\midrule
\textbf{Completed Stages}$\boldsymbol{\uparrow}$ & & $2.44$ & $0.91$ & $2.26$ & $1.7$ & \bestgroupone{$2.81$} & \bestgrouptwo{$2.69$}\\
\midrule
\textbf{Drop}$\boldsymbol{\downarrow}$ & & $-$ & $1.53$ & $-$ & $0.56$ & $-$ & \bestgrouptwo{$\mathbf{0.12}$}\\
\bottomrule
\end{tabular}
}
    \vspace{-0.3em}
    \captionof{table}{
        Average completed stages in default and variant environments. Max denotes total stages per task. Best results in the \textbf{\textcolor{groupone}{Default}} and \textbf{\textcolor{grouptwo}{Variant}} settings are highlighted in \textbf{\textcolor{groupone}{blue}} and \textbf{\textcolor{grouptwo}{green}}, respectively.}
    \label{table:real_robot_result}
\end{minipage}
\vspace{-0.6em}
\end{figure*}

\textbf{Setup}. We evaluate our methods on four challenging real-world manipulation tasks: Drawer, Insertion, Pick Place and Cup Stack. An overview of the setup is shown in Figure~\ref{table:real_robot_result}. Similar to simulation, our vision system uses three cameras: left-, right-, and gripper-views. To evaluate robustness to camera perturbations, we additionally introduce a right-side camera with a substantially different pose. The default environment uses the left camera, default right camera, and gripper camera, while the variant environment replaces the default right camera with the perturbed one.

\textbf{Dataset}. For collecting demonstrations, we use teleoperation system consisting of a leader robot and a follower robot. For Drawer, Pick Place and Cup Stack three tasks, we collected 50 demonstrations per task with language instruction. For Insertion task, we have collected 100 language-conditioned trajectories. The details of our tasks are in the Appendix~\ref{subsec:task_description}.

\textbf{Metrics}. Given the long-horizon nature of the tasks and the presence of camera perturbations, we adopt a structured scoring metric for fair and fine-grained evaluation. Each task is decomposed into multiple stages, with each completed stage contributing one point. The final score is the average sum of completed intermediate stages. Additional details are provided in Appendix~\ref{subsec:task_metric}.

\textbf{Results}. We compare RayViT and RayViT-cls against the RGB variant on four challenging real-world tasks, evaluating each model with 20 rollouts per task. As shown in Table~\ref{table:real_robot_result}, our approaches maintain strong performance in real-world settings. Under the default camera configuration, both methods achieve performance comparable to the RGB baseline. Under the variant setting, however, our methods exhibit clear robustness advantages, improving the accumulated score by at least 0.8 points and demonstrating substantially improved resilience to camera perturbations.

Interestingly, RayViT achieves the strongest performance under both default and variant camera settings, surpassing RayViT-cls. One possible explanation is that the real-world images are collected at twice the resolution of the simulation data, resulting in longer image patch sequences. In this setting, injecting richer geometric information may provide greater benefits.


\section{Limitations and Future Work}
\label{sec:limitations}

Our current work has three main limitations. First, we focus on injecting camera geometry into pretrained ViT-based visual encoders, while powerful convolution-based visual models are not considered. Second, our method is developed in a multi-view setting and relies on a task-relevant anchor view for the cross-view consistency objective.

Several directions may address these limitations. A natural extension is to develop geometry-conditioning mechanisms for pretrained convolutional visual backbones. We also plan to investigate the behavior of our method under varying numbers of viewpoints, including both sparse-view and dense-view settings. More broadly, we believe a promising direction is to pretrain geometry-conditioned visual encoders from the outset, enabling stronger alignment between geometric priors and learned visual representations for robust robotic policy learning.


\section{Conclusion}
\label{sec:conclusion}

We presented RayViT, a lightweight framework for injecting camera geometry into pretrained ViT-based visual encoders to improve the robustness of imitation learning policies under camera viewpoint perturbations. Our approach adopts two geometry injection schemes: a ray-map–conditioned class token that replaces the standard CLS token with a geometry-aware scene descriptor, and ray-based positional embeddings added alongside RoPE to provide explicit 3D grounding cues. We further introduce a cross-view consistency objective, finding that geometric grounding may serve as a prerequisite for effective auxiliary representation learning in multi-view settings. Extensive experiments on RoboCasa and real-world manipulation tasks demonstrate that RayViT significantly improves robustness to camera perturbations. Ablation studies show that both ray-map conditioning and the consistency objective contribute independently, with benefits persisting across pretrained and randomly initialized backbones, as well as different pretrained ViT encoders.



\clearpage
\acknowledgments{If a paper is accepted, the final camera-ready version will (and probably should) include acknowledgments. All acknowledgments go at the end of the paper, including thanks to reviewers who gave useful comments, to colleagues who contributed to the ideas, and to funding agencies and corporate sponsors that provided financial support.}


\bibliography{example}  

\clearpage

\appendix
\section{Score-based Diffusion Policy}
Diffusion models are generative models that learn to generate new samples through learning to reverse a Gaussian Perturbation process. In this work, we apply a score-based diffusion model to formulate the policy representation $\pi_\theta(\bar{a}\mid s)$. This perturbation and its inverse process can be expressed through a Stochastic Differential Equation (SDE):

\begin{equation}
d\bar{a} = \left(\beta_t \sigma_t - \dot{\sigma}_t\right)\sigma_t \nabla_a \log p_t(\bar{a}\mid s)\, dt + \sqrt{2\beta_t \sigma_t}\, d\omega_t ,
\end{equation}

where $\beta_t$ determines the noise injection rate, $d\omega_t$ represents infinitesimal Gaussian noise, and \mbox{$p_t(\bar{a}\mid s)$} denotes the score function of the diffusion process. 


\section{Simulation Experiment Details}
    \subsection{Baseline Introduction}
    \label{subsec: baseline introduction}
    We mainly selected four representative baselines:
    1) \textbf{Adapt3R} combines pretrained 2D semantic features with 3D geometric information, using 3D points primarily to localize semantic cues rather than to directly encode full scene structure. 
    2) \textbf{PMP-xyz} and \textbf{PMP} use explicit 3D point-map representations, treating point maps as image-like inputs that preserve dense geometric information. PMP-xyz uses only 3D point-map features, while PMP concatenates 3D tokens with 2D semantic tokens extracted from RGB images. 
    3) \textbf{CamPose} also incorporates ray maps for imitation learning, but treats them as an additional input modality and fuses them with semantic tokens through late concatenation. 
    4) \textbf{RGB} is our image-only variant, which removes ray-map conditioning and passes RGB observations directly through DINOv3.

    \subsection{PMP Result}
    In the main results reported in Table~\ref{table:main_result}, we include PMP-xyz, which uses only 3D point-map features. The original PMP paper also introduces a variant, referred to as PMP, that combines point-map tokens with image tokens. Table~\ref{table:pmp_result} provides the corresponding comparison with this PMP variant.
\begin{table*}[h]
\begin{center}
\resizebox{\textwidth}{!}{%
\renewcommand{\arraystretch}{1.3}
\setlength{\aboverulesep}{0pt}
\setlength{\belowrulesep}{0pt}
\begin{tabular}{l|cc|cc|cc|cc}
\toprule
\multirow{2}{*}{\textbf{Task}} & \multicolumn{2}{c|}{PMP} & \multicolumn{2}{c|}{PMP-xyz} & \multicolumn{2}{c|}{\textbf{RayVit-cls}} & \multicolumn{2}{c}{\textbf{RayVit}}\\

\cmidrule(lr){2-3}\cmidrule(lr){4-5}\cmidrule(lr){6-7}\cmidrule(lr){8-9}
& Default & Variant & Default & Variant & Default & Variant & Default & Variant\\
\midrule

PnPCounterToMicrowave
& $2.7 \scriptstyle \pm 1.2$
& $0.0 \scriptstyle \pm 0.0$
& $2.0 \scriptstyle \pm 0.0$
& $0.0 \scriptstyle \pm 0.0$
& $1.3 \scriptstyle \pm 0.9$
& $0.0 \scriptstyle \pm 0.0$
& $2.0 \scriptstyle \pm 1.6$
& $\mathbf{2.0 \scriptstyle \pm 1.6}$

\\
PnPCounterToSink
& $0.0 \scriptstyle \pm 0.0$
& $0.0 \scriptstyle \pm 0.0$
& $2.0 \scriptstyle \pm 2.0$
& $2.0 \scriptstyle \pm 0.0$
& $0.7 \scriptstyle \pm 0.9$
& $1.3 \scriptstyle \pm 1.9$
& $\mathbf{2.7 \scriptstyle \pm 0.9}$
& $\mathbf{1.3 \scriptstyle \pm 1.9}$

\\
PnPMicrowaveToCounter
& $\mathbf{16.0 \scriptstyle \pm 6.0}$
& $1.3 \scriptstyle \pm 2.3$
& $10.0 \scriptstyle \pm 2.0$
& $2.7 \scriptstyle \pm 3.1$
& $13.3 \scriptstyle \pm 0.9$
& $\mathbf{8.7 \scriptstyle \pm 3.4}$
& $6.7 \scriptstyle \pm 2.5$
& $6.7 \scriptstyle \pm 3.4$

\\
PnPSinkToCounter
& $9.3 \scriptstyle \pm 3.1$
& $2.7 \scriptstyle \pm 3.1$
& $4.7 \scriptstyle \pm 1.2$
& $2.0 \scriptstyle \pm 2.0$
& $13.3 \scriptstyle \pm 5.2$
& $\mathbf{12.0 \scriptstyle \pm 4.3}$
& $9.3 \scriptstyle \pm 5.0$
& $6.0 \scriptstyle \pm 3.3$

\\
\midrule
OpenDrawer
& $44.0 \scriptstyle \pm 5.3$
& $18.0 \scriptstyle \pm 24.6$
& $\mathbf{49.3 \scriptstyle \pm 1.2}$
& $18.0 \scriptstyle \pm 7.2$
& $44.7 \scriptstyle \pm 6.2$
& $\mathbf{40.7 \scriptstyle \pm 5.0}$
& $34.7 \scriptstyle \pm 6.8$
& $29.3 \scriptstyle \pm 5.0$

\\
CloseDrawer
& $85.3 \scriptstyle \pm 3.1$
& $34.7 \scriptstyle \pm 47.9$
& $\mathbf{99.3 \scriptstyle \pm 1.2}$
& $55.3 \scriptstyle \pm 23.0$
& $87.3 \scriptstyle \pm 2.5$
& $81.3 \scriptstyle \pm 4.1$
& $81.3 \scriptstyle \pm 9.4$
& $\mathbf{84.7 \scriptstyle \pm 4.1}$
\\
\midrule

TurnOnStove
& $18.7 \scriptstyle \pm 3.1$
& $10.7 \scriptstyle \pm 15.0$
& $34.7 \scriptstyle \pm 6.4$
& $8.7 \scriptstyle \pm 1.2$
& $28.0 \scriptstyle \pm 0.0$
& $24.0 \scriptstyle \pm 5.9$
& $28.7 \scriptstyle \pm 1.9$
& $\mathbf{30.7 \scriptstyle \pm 2.5}$
\\
TurnOffStove
& $12.0 \scriptstyle \pm 4.0$
& $3.3 \scriptstyle \pm 4.2$
& $16.0 \scriptstyle \pm 2.0$
& $7.3 \scriptstyle \pm 9.2$
& $14.7 \scriptstyle \pm 2.5$
& $13.3 \scriptstyle \pm 2.5$
& $14.0 \scriptstyle \pm 1.6$
& $\mathbf{17.3 \scriptstyle \pm 0.9}$
\\
\midrule

TurnOnSinkFaucet
& $50.7 \scriptstyle \pm 5.0$
& $31.3 \scriptstyle \pm 23.2$
& $68.7 \scriptstyle \pm 2.3$
& $33.3 \scriptstyle \pm 10.1$
& $\mathbf{70.0 \scriptstyle \pm 3.3}$
& $\mathbf{68.0 \scriptstyle \pm 9.9}$
& $65.3 \scriptstyle \pm 2.5$
& $66.0 \scriptstyle \pm 4.9$
\\
TurnOffSinkFaucet
& $48.7 \scriptstyle \pm 1.2$
& $23.3 \scriptstyle \pm 20.0$
& $\mathbf{76.0 \scriptstyle \pm 6.9}$
& $36.7 \scriptstyle \pm 1.2$
& $52.7 \scriptstyle \pm 5.2$
& $48.7 \scriptstyle \pm 5.2$
& $56.7 \scriptstyle \pm 4.1$
& $\mathbf{51.3 \scriptstyle \pm 9.0}$
\\
TurnSinkSpout
& $30.7 \scriptstyle \pm 3.1$
& $28.7 \scriptstyle \pm 6.1$
& $\mathbf{72.0 \scriptstyle \pm 2.0}$
& $\mathbf{48.0 \scriptstyle \pm 6.9}$
& $40.7 \scriptstyle \pm 7.4$
& $\mathbf{43.3 \scriptstyle \pm 3.4}$
& $42.7 \scriptstyle \pm 3.8$
& $43.3 \scriptstyle \pm 10.5$
\\
\midrule

CoffeePressButton
& $74.0 \scriptstyle \pm 15.9$
& $36.0 \scriptstyle \pm 45.1$
& $75.3 \scriptstyle \pm 8.1$
& $53.3 \scriptstyle \pm 8.1$
& $\mathbf{92.0 \scriptstyle \pm 4.3}$
& $\mathbf{85.3 \scriptstyle \pm 5.0}$
& $84.7 \scriptstyle \pm 4.7$
& $84.7 \scriptstyle \pm 2.5$
\\
TurnOnMicrowave
& $55.3 \scriptstyle \pm 6.4$
& $23.3 \scriptstyle \pm 40.4$
& $46.0 \scriptstyle \pm 8.7$
& $10.7 \scriptstyle \pm 5.0$
& $72.0 \scriptstyle \pm 5.7$
& $48.0 \scriptstyle \pm 7.1$
& $64.0 \scriptstyle \pm 4.3$
& $\mathbf{54.0 \scriptstyle \pm 2.8}$
\\
TurnOffMicrowave
& $70.7 \scriptstyle \pm 1.2$
& $34.0 \scriptstyle \pm 42.0$
& $50.7 \scriptstyle \pm 9.0$
& $7.3 \scriptstyle \pm 4.6$
& $\mathbf{91.3 \scriptstyle \pm 2.5}$
& $\mathbf{62.7 \scriptstyle \pm 12.0}$
& $85.3 \scriptstyle \pm 2.5$
& $61.3 \scriptstyle \pm 10.9$
\\
\midrule

CoffeeServeMug
& $39.3 \scriptstyle \pm 2.3$
& $16.7 \scriptstyle \pm 22.0$
& $\mathbf{52.7 \scriptstyle \pm 8.1}$
& $33.3 \scriptstyle \pm 6.4$
& $\mathbf{52.0 \scriptstyle \pm 3.3}$
& $42.0 \scriptstyle \pm 9.9$
& $50.7 \scriptstyle \pm 2.5$
& $\mathbf{46.7 \scriptstyle \pm 4.7}$
\\
CoffeeSetupMug
& $12.7 \scriptstyle \pm 1.2$
& $7.3 \scriptstyle \pm 9.2$
& $14.7 \scriptstyle \pm 1.2$
& $14.0 \scriptstyle \pm 2.0$
& $\mathbf{16.7 \scriptstyle \pm 4.7}$
& $\mathbf{12.7 \scriptstyle \pm 1.9}$
& $7.3 \scriptstyle \pm 3.4$
& $10.7 \scriptstyle \pm 1.9$
\\
\midrule
{\textbf{Average Success Rate}}
& $35.6 $
& $17.0 $
& $42.1 $
& $20.8 $
& $\mathbf{43.2}$
& $\mathbf{37.0}$
& $39.8$
& $\mathbf{37.3}$
\\
\midrule

{\textbf{Performance Drop}}

& $-$
& $18.6$
& $-$
& $21.3 $
& $-$
& $\mathbf{6.2}$
& $-$
& $\mathbf{2.5}$
\\
\bottomrule
\end{tabular}
}
\end{center}
\caption{Comparison of our methods with PMP~\cite{jia2025pointmappolicy} on RoboCasa~\cite{robocasa2024}. All models were trained for 50 epochs using 50 human demonstrations per task and evaluated over 50 episodes per task.}
\label{table:pmp_result}
\end{table*}


\section{Real Robot Experiment Details}
\begin{figure}[h]    
    \centering
    \includegraphics[width=0.95\linewidth]{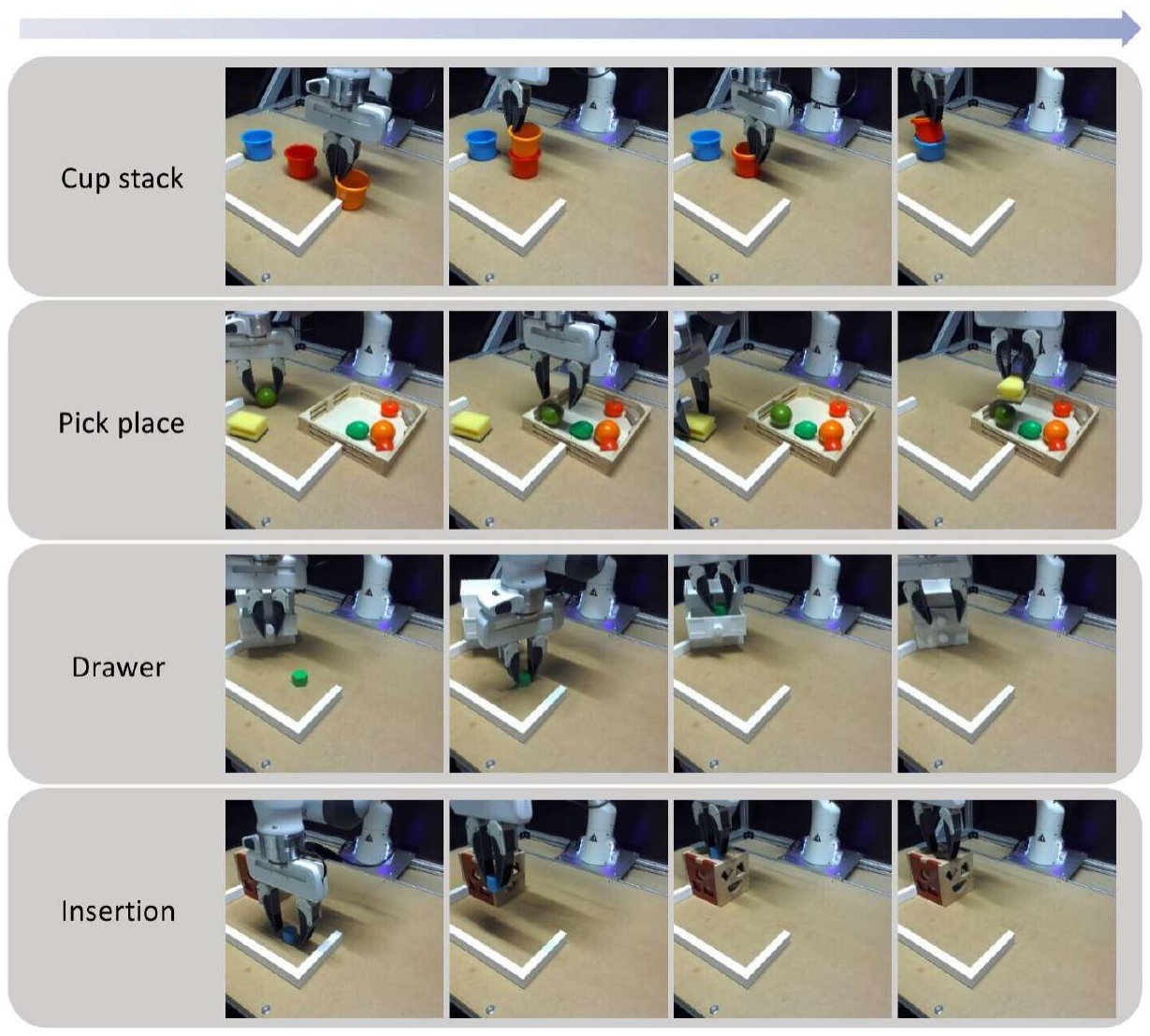}
    \caption{Overview of real robot tasks}
    \label{fig:real_robot_task}
\end{figure}

    We conduct four real-world experiments on a Franka Panda Robot: Cup Stack, Pick Place, Drawer and Insertion. Table~\ref{fig:real_robot_task} shows an overview of our real-world tasks.

    \subsection{Setup}
        The external viewpoints in our real world experiments are captured using ZED Mini stereo cameras, and the gripper view uses an Intel RealSense D405. We use the left-eye image from each stereo camera and resize all observations to 256×256. Experiments are conducted on a 7-DOF Franka Panda robot with an 8-dimensional action space comprising joint positions and gripper state.

    \subsection{Task Metric}
    \label{subsec:task_metric}
        Given the long-horizon nature of the tasks and the presence of camera perturbations, we decompose each task into several discrete stages. The final score is computed as the total number of successfully completed stages. Details of the scoring metric design are provided in Table~\ref{table:task_metrics}.

        \begin{table}[h]
    \centering
     \resizebox{\textwidth}{!}{
    \begin{tabular}{c|c|c}
         \toprule
        \textbf{Cup Stack}                    & \textbf{Pick Place}         & \textbf{Score} \\
        \midrule
        pick up the orange cup                & pick up the pear or sponge  & 1 \\
        \midrule
        stack the orange cup onto the red cup & place it onto the tray      & 2 \\
        \midrule
        pick up the stacked two cups          & pick up the rest object on the table  & 3 \\
        \midrule
        stack them onto the blue cup          & place it onto the tray      & 4 \\
        \midrule

        \addlinespace[0.75em]

        \midrule
        \textbf{Drawer}         & \textbf{Insertion}                          & \textbf{Score} \\
        \midrule
        open the upper drawer   & pick up the cylinder                        & 1 \\
        \midrule
        pick up the green prism & release the cylinder above the matching hole & 2 \\
        \midrule
        place the prism inside the drawer & insert the cylinder successfully  & 3 \\
        \midrule
        close the drawer        &          -                                  & 4 \\
        \bottomrule
        
    \end{tabular}
    }
    \vspace{0.15cm}
    \caption{Task score metric details of real robot and evaluation standards.}
    \label{table:task_metrics}
\end{table}

    \subsection{Task Description}
    \label{subsec:task_description}
        \textbf{Cup Stack}. In the cup stacking task, the robot is presented with three cups of different colors and sizes that must be stacked according to size. The large blue cup is positioned on the left, the medium red cup in the center, and the small orange cup on the right. The robot must first pick up the orange cup and place it onto the red cup, then lift the resulting orange–red stack and place it onto the blue cup. The horizontal positions of the cups are fixed, while their vertical positions vary.

        \textbf{Pick Place}. In the pick-and-place task, a wooden tray containing five unrelated toys is placed on the table. A toy pear and a sponge are positioned to the left of the tray. The robot’s objective is to pick up the pear and the sponge and place them into the tray. The positions of the pear, sponge, and toys inside the tray may vary, while the position and orientation of the tray remain fixed. During data collection, the order in which the pear and sponge are manipulated is randomized.

        \textbf{Drawer}. In the drawer task, a two-layer white drawer is placed on the left side of the workspace, and a green prism is positioned on the right. The robot must open the upper drawer, pick up the green prism and place it inside the drawer, and then close the drawer. The position and orientation of the drawer are fixed, whereas the position of the prism vary.

        \textbf{Insertion}. In the insertion task, a wooden cube-shaped box with differently shaped holes on its faces is placed on the left side of the workspace. The top face contains circular, triangular, and hexagonal holes. A blue cylinder is positioned on the right side. The robot’s objective is to pick up the cylinder and insert it through the circular hole such that it falls vertically into the box. The position and orientation of the box are fixed, while the position of the cylinder vary.

\section{Hyper Parameters}
    \subsection{Policy Hyper Parameters}
    
\begin{table*}[ht]
\centering
\large
\renewcommand{\arraystretch}{1.15}
\setlength{\tabcolsep}{16pt}
\begin{tabular}{l|c}
\toprule
\textbf{Hyperparameter} &  decoder-only x-LSTM\\
\hline
Number of x-Blocks & 8  \\
Attention Heads & 8 \\
Action Chunk Size & 10 \\
Embedding Dimension & 768 \\
Goal Lang Encoder & CLIP ViT-B/32 \\
Optimizer & AdamW \\
Betas & [0.9, 0.95] \\
Learning Rate & 1e-4 \\
Weight Decay & 0.05 \\
Batch Size & 256 \\
$\sigma_{\text{max}}$ & 80 \\
$\sigma_{\text{min}}$ & 0.001 \\
$\sigma_t$ & 0.5 \\
EMA & True \\
Time steps & Exponential \\
Sampler & DDIM \\
Sampling Step & 4 \\
\bottomrule
\end{tabular}
\caption{Hyperparameters Summary of Policy.}
\label{tab:hyperparameters}
\end{table*}
    Where, $\sigma_{\text{max}}$ is the starting noise level, $\sigma_{\text{min}}$ is the floor of the noise schedule and $\sigma_{t}$ is standard deviation of the target data.

    \subsection{RayViT Hyper Parameters}
    
\begin{table*}[ht]
\centering
\large
\renewcommand{\arraystretch}{1.15}
\setlength{\tabcolsep}{16pt}
\begin{tabular}{l|c}
\toprule
\textbf{Hyperparameter} &  RayViT\\
\hline
Number of Attention Blocks & 2  \\
Number of Attention Heads & 4 \\
Mlp Ratio & 4 \\
Embedding Dimension & 384 \\
qkv\_bias & True \\
proj\_bias & True \\
ffn\_bias & True \\
Cosine Similarity Layer & 10th \\
Weight of Cosine Similarity Loss & 0.01 \\
Number of Ray patch projection & 0.15M \\
Number of Gated Attention Module & 3.85M \\
Number of DINOv3 ViT\_small & 21.6M \\
\bottomrule
\end{tabular}
\caption{Hyperparameters Summary of RayViT.}
\label{tab:rayvit_hyperparameters}
\end{table*}

\section{Compute Resources}

    In both simulation and real-world experiments, RayViT uses DINOv3 ViT-Small as the primary visual backbone. For simulation tasks, models are trained for 50 epochs on a single NVIDIA A100-SXM4-40GB GPU with a batch size of 256, finishing within 8 hours.  
    For real-world experiments, models are trained for 60 epochs on a single GeForce RTX 5090 32GB GPU with a batch size of 64 and 4 gradient accumulation steps, finishing within 7 hours.

\end{document}